\documentclass{article}

\PassOptionsToPackage{numbers, compress}{natbib}

\usepackage[preprint]{neurips_2019}

\usepackage[utf8]{inputenc} 
\usepackage[T1]{fontenc}    
\usepackage{hyperref}       
\usepackage{url}            
\usepackage{booktabs}       
\usepackage{amsfonts}       
\usepackage{nicefrac}       
\usepackage{microtype}      

\usepackage{epsfig}
\usepackage{graphics}
\usepackage{multirow}
\usepackage[dvipsnames]{xcolor}

\newcommand*{\affaddr}[1]{#1} 
\newcommand*{\affmark}[1][*]{\textsuperscript{#1}}
\newcommand*{\email}[1]{\texttt{#1}}

\title{\large Contextual Grounding of Natural Language Entities in Images}

\author{%
Farley Lai\affmark[1], Ning Xie\affmark[2]\thanks{Work performed as a NEC Labs intern}, Derek Doran\affmark[2], Asim Kadav\affmark[1]\\
\affaddr{\affmark[1]Machine Learning Department, NEC Laboratories America, Inc.}\\
\affaddr{\affmark[2]Department of Computer Science, Wright State University}\\
\email{farleylai@nec-labs.com}\\
\email{\{xie.25, derek.doran\}@wright.edu}\\
\email{asim@nec-labs.com}\\
}

\begin{document}

\maketitle

\begin{abstract}
In this paper, we introduce a contextual grounding approach that captures the context in corresponding text entities and image regions to improve the grounding accuracy.
Specifically, the proposed architecture accepts pre-trained text token embeddings and image object features from an off-the-shelf object detector as input.
Additional encoding to capture the positional and spatial information can be added to enhance the feature quality.
There are separate text and image branches facilitating respective architectural refinements for different modalities.
The text branch is pre-trained on a large-scale masked language modeling task while the image branch is trained from scratch.
Next, the model learns the contextual representations of the text tokens and image objects through layers of high-order interaction respectively.
The final grounding head ranks the correspondence between the textual and visual representations through cross-modal interaction.
In the evaluation, we show that our model achieves the state-of-the-art grounding accuracy of 71.36\% over the Flickr30K Entities dataset.
No additional pre-training is necessary to deliver competitive results compared with related work that often requires task-agnostic and task-specific pre-training on cross-modal dadasets. The implementation is publicly available at \url{https://gitlab.com/necla-ml/grounding}.
\end{abstract}

\section{Introduction}
\label{sec:intro}

\begin{figure}[ht]
  \centering
  \includegraphics[width=0.5\linewidth]{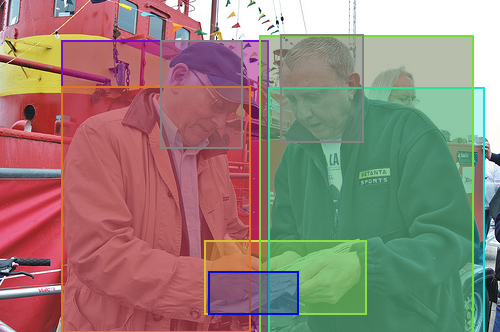}
  \caption{Example image from Flickr30K Entities annotated with bounding boxes corresponding to entities in the caption "\textcolor{red}{A man} wearing \textcolor{BurntOrange}{a tan coat} signs \textcolor{GreenYellow}{papers} for \textcolor{LimeGreen}{another man} wearing \textcolor{Cyan}{a blue coat}."}
  \label{fig:example}
\end{figure}

Cross-modal reasoning is challenging for grounding entities and objects in different modalities.
Representative tasks include visual question answering (VQA) and image captioning that leverage grounded features between text and images to make predictions.
While recent advances in these tasks achieve impressive results, the quality of the correspondence between textual entities and visual objects in both modalities is not necessarily convincing or interpretable \citep{Liu:2017ve}.
This is likely because the grounding from one modality to the other is trained implicitly and the intermediate results are not often evaluated as explicitly as in object detection.
To address this issue, \citet{Plummer:2015ve} created the \emph{Flickr30K Entities} dataset with precise annotations of the correspondence between language phrases and image regions to ease the evaluation of visual grounding.
In Figure\ref{fig:example}, two men are referred to in the caption as separate entities.
To uniquely ground the two men in the image, the grounding algorithm must take respective context and attributes into consideration for learning the correspondence.

Over the years, many deep learning based approaches were proposed to tackle this localization challenge.
The basic idea is to derive representative features for each entity as well as object, and then score their correspondence.
In the modality of caption input, individual token representations usually start with the word embeddings followed by a recurrent neural network (RNN), usually Long Short-Term Memory (LSTM) or Gated Recurrent Units (GRU),  to capture the contextual meaning of the text entity in a sentence.
On the other hand, the visual objects in image regions of interest (RoI) are extracted through object detection.
Each detected object typically captures limited context through the receptive fields of 2D convolutions.
Advanced techniques such as the \emph{feature pyramid network} (FPN) \citep{Lin:2017vv} enhance the representations by combining features at different semantic levels w.r.t. the object size.
Even so, those conventional approaches are limited to extracting relevant long range context in both text and images effectively.
In view of this limitation, non-local attention techniques were proposed to address the long range dependencies in natural language processing (NLP) and computer vision (CV) tasks \citep{Vaswani:2017ul, Wang:2018ut}.
Inspired by this advancement, we introduce the contextual grounding approach to improving the representations through extensive intra- and inter-modal interaction to infer the contextual correspondence between text entities and visual objects.


\paragraph{Related Work.} \label{sec:related}
On the methodology of feature interaction, the \emph{Transformer} architecture \citep{Vaswani:2017ul} for machine translation demonstrates a systematic approach to efficiently computing the interaction between language elements.
Around the same time, \emph{non-local networks} \citep{Wang:2018ut} generalize the transformer to the CV domain, supporting feature interaction at different levels of granularity from feature maps to pooled objects.
Recently, the image transformer \citep{Parmar:2018vc} adapts the original transformer architecture to the image generation domain by encoding spatial information in pixel positions while we deal with image input at the RoI level for grounding.
The following work in \citep{Devlin:2018uk} proposed BERT as a pre-trained transformer encoder on large-scale masked language modeling, facilitating training downstream tasks to achieve state-of-the-art (SOTA) results.
Our work extends BERT to the cross-modal grounding task by jointly learning contextual representations of language entities and visual objects.
Coincidentally, another line of work named \emph{VisualBERT} \citep{Li:2019wy} also integrates BERT to deal with grounding in a single transformer architecture.
However, their model requires both task-agnostic and task-specific pre-training on cross-modal datasets to achieve competitive results.
Ours, on the contrary, achieves SOTA results without additional pre-training and allows respective architectural concerns for different modalities.

\section{Contextual Grounding}
\label{sec:design}

\begin{figure}[ht]
  \centering
  \includegraphics[width=\linewidth]{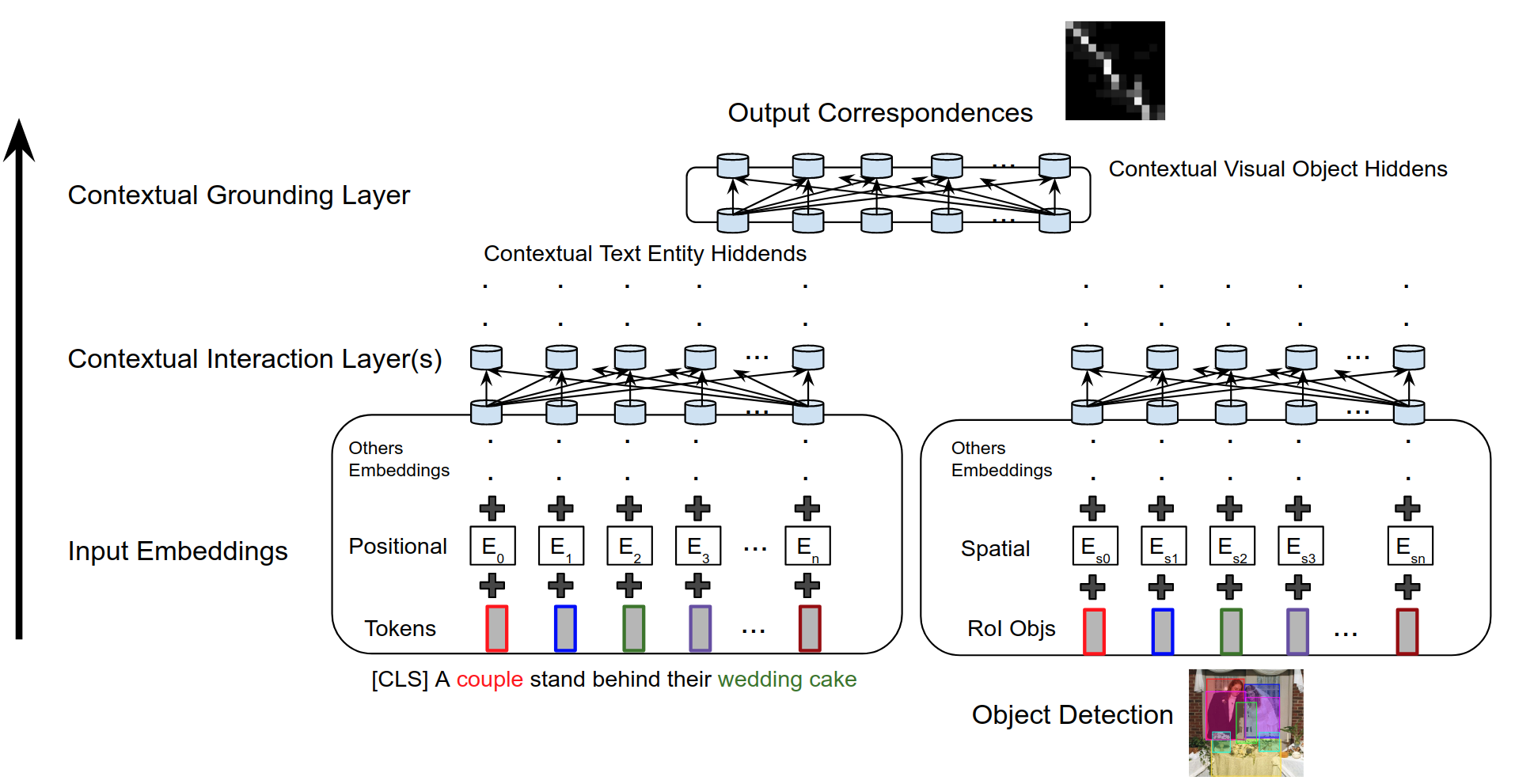}
  \caption{Contextual grounding architecture}
  \label{fig:arch}
\end{figure}

The main approach of previous work is to use RNN/LSTM to extract high level phrase representations and then apply different attention mechanisms to rank the correspondence to visual regions.
While the hidden representations of the entity phrases take the language context into consideration, the image context around visual objects is in contrast limited to object detection through 2D receptive fields.
Nonetheless, there is no positional ordering as in text for objects in an image to go through the RNN to capture potentially far apart contextual dependencies.
In view of the recent advances in NLP, the transformer architecture proposed by \citep{Vaswani:2017ul} addresses the long range dependency through pure attention techniques.
Without RNN being incorporated, the transformer enables text tokens to efficiently interact with each other pairwise regardless of the range.
The ordering information is injected through additional positional encoding.
Enlightened by this breakthrough, the corresponding contextual representations of image RoIs may be derived through intra-modal interaction with encoded spatial information.
We hypothesize that the grounding objective would guide the attention to the corresponding context in both the text and image with improved accuracy.
Consequently, we propose the contextual grounding architecture as shown in Figure~\ref{fig:arch}.
The model is composed of two transformer encoder branches for both text and image inputs to generate their respective contextual representations for the grounding head to decide the correspondence.
The text branch is pre-trained from the BERT base model \citep{Devlin:2018uk} which trains a different positional embedding from the original transformer \citep{Vaswani:2017ul}.
On the other hand, the image branch takes RoI features as input objects from an object detector.
Correspondingly, we train a two layer MLP to generate the spatial embedding given the absolute spatial information of the RoI location and size normalized to the entire image.
Both branches add the positional and spatial embedding to the tokens and RoIs respectively as input to the first interaction layer.
At each layer, each hidden representation performs self-attention to each other to generate a new hidden representation as layer output.
The self-attention may be multi-headed to enhance the representativeness as described in \citep{Vaswani:2017ul}.
At the end of each branch, the final hidden state is fed into the grounding head to perform the cross-modal attention with text entity hidden states as queries and image object hidden representations as the keys.
The attention responses serve as the matching correspondences.
If the correspondence does not match the ground truth, the mean binary cross entropy loss per entity is back propagated to guide the interaction across the branches.
We evaluate the grounding recall on the Flickr30K Entities dataset and compare the results with SOTA work in the next section.

\section{Evaluation}
\label{sec:evaluation}

\begin{table}[ht]
\newcommand{\tabincell}[2]{\begin{tabular}{@{}#1@{}}#2\end{tabular}}
\begin{tabular}{lccccc}
\toprule
\textbf{Model}          & \textbf{Detector}                 & \textbf{R@1}  & \textbf{R@5}  & \textbf{R@10} & \textbf{Upper Bound}  \\
\midrule
\citet{Plummer:2015ve}  & Fast RCNN                         & 50.89             & 71.09             & 75.73             & 85.12             \\
\citet{Yeh:2017td}      & YOLOv2                            & 53.97             & -                 & -                 & -                 \\
\citet{Hinami:2017uz}   & Query-Adaptive RCNN               & 65.21             & -                 & -                 & -                 \\
BAN \citep{Kim:2018wu}  & Bottom-Up \citep{Anderson:2018ue} & 69.69             & 84.22             & 86.35             & 87.45             \\
\midrule
Ours L1-H2-abs          & Bottom-Up \citep{Anderson:2018ue} & \textbf{71.36}    & 84.76             & 86.49             & 87.45             \\
\midrule
Ours L1-H1-abs          & Bottom-Up \citep{Anderson:2018ue} & 71.21             & \textbf{84.84}    & \textbf{86.51}    & 87.45             \\
Ours L1-H1              & Bottom-Up \citep{Anderson:2018ue} & 70.75             & 84.75             & 86.39             & 87.45             \\
Ours L3-H2-abs          & Bottom-Up \citep{Anderson:2018ue} & 70.82             & 84.59             & 86.49             & 87.45             \\
Ours L3-H2              & Bottom-Up \citep{Anderson:2018ue} & 70.39             & 84.68             & 86.35             & 87.45             \\
Ours L6-H4-abs          & Bottom-Up \citep{Anderson:2018ue} & 69.71             & 84.10             & 86.33             & 87.45             \\
\bottomrule
\\
\end{tabular}
\caption{Accuracy on Flickr30K Entities test split where $L$, $H$ and $abs$ denote the number of layers, attention heads and whether the absolute spatial embedding is employed in the image branch.}
\label{table:performance}
\end{table}

\begin{table}[ht]
\newcommand{\tabincell}[2]{\begin{tabular}{@{}#1@{}}#2\end{tabular}}
\makebox[\textwidth][c]{
\begin{tabular}{lllllllll}
\toprule
\textbf{Model}          &\textbf{People}    &\textbf{Clothing}  &\textbf{Body Parts}    &\textbf{Animals}   &\textbf{Vehicles}  &\textbf{Instruments}   &\textbf{Scene}         &\textbf{Other} \\
\midrule
\citet{Plummer:2015ve}  &64.73              &46.88              &17.21                  &65.83              &68.75              &37.65                  &51.39                  &31.77          \\
\citet{Yeh:2017td}      &68.71              &46.83              &19.50                  &70.07              &73.75              &39.50                  &60.38                  &32.45          \\
\citet{Hinami:2017uz}   &78.17              &61.99              &35.25                  &74.41              &76.16              &\textbf{56.69}         &68.07                  &47.42          \\
BAN \citep{Kim:2018wu}      &79.90              &74.95              &\textbf{47.23}         &81.85              &76.92              &43.00                  &68.69                  &51.33          \\
\midrule
Ours L1-H2-abs          &\textbf{81.95} &\textbf{76.5}  &46.27              &\textbf{82.05} &\textbf{79.0}  &35.8           &\textbf{70.23} &\textbf{53.53} \\
\midrule
\# of instances              &5656           &2306           &523                &518            &400            &162            &1619           &3374      \\
\bottomrule
\\
\end{tabular}
}
\caption{Per Flickr30k entity type recall(\%) breakdown.}
\label{table:breakdown}
\end{table}

Our contextual grounding approach uses the transformer encoder to capture the context in both text entities and image objects.
While the text branch is pre-trained from BERT \citep{Devlin:2018uk}, the image branch is trained from scratch.
In view of the complexity of the transformer, previous work \citep{Girdhar:2019uo} has shown the performance varies with different numbers of interaction layers and attention heads.
Also, the intra-modal object interaction does not necessarily consider the relationship in space unless some positional or spatial encoding is applied.
In our evaluation, we vary both the number of layers and heads, along with adding the spatial encoding to explore the performance variations summarized in Table~\ref{table:performance}.
We achieve the SOTA results in all top 1, 5 and 10 recalls based on the same object detector as used by previous SOTA BAN \citep{Kim:2018wu}.
The breakdown of per entity type recalls is given in Table~\ref{table:breakdown}.
Six out of the eight entity type recalls benefit from our contextual grounding.
Interestingly, the recall of the instrument type suffers.
This may be due to the relative small number of instrument instances in the dataset preventing the model from learning the context well.
On the other hand, compared with the text branch consisting of $12$ layers and $12$ heads with hidden size of $768$ dimensions, the best performance is achieved with the image branch having $1$ layer, $2$ attention heads and hidden size of $2048$ dimensions.
Moreover, adding the spatial embedding consistently improves the accuracy by $0.5\%$ or so.
This is likely because image objects, unlike word embedding requiring the context to produce representative hidden states for its meaning, may already capture some neighborhood information through receptive fields.

\begin{table}[ht]
\newcommand{\tabincell}[2]{\begin{tabular}{@{}#1@{}}#2\end{tabular}}
\makebox[\textwidth][c]{
\begin{tabular}{lcccccccc}
\toprule
\multirow{2}{*}{\textbf{Model}}     &\multicolumn{2}{c}{\textbf{R@1}}   &\multicolumn{2}{c}{\textbf{R@5}}   &\multicolumn{2}{c}{\textbf{R@10}}  &\multicolumn{2}{c}{\textbf{Upper Bound}}   \\
                                    &\textbf{Dev}    &\textbf{Test}     &\textbf{Dev}    &\textbf{Test}     &\textbf{Dev}    &\textbf{Test}     &\textbf{Dev}    &\textbf{Test}             \\
\midrule                                     
VisualBERT w/o COCO Pre-training     &68.07  &-                          &83.98  &-                          &86.24  &-                          &86.97  &87.45                              \\
VisualBERT \citep{Li:2019wy}        &70.40  &71.33                      &84.49  &\textbf{84.98}             &86.31  &\textbf{86.51}             &       &                                   \\
\midrule
Ours L1-H2-abs                      &69.8   &\textbf{71.36}             &84.22  &84.76                      &86.21  &86.49                      &86.97  &87.45                              \\
\bottomrule
\\
\end{tabular}
}
\caption{Comparison with VisualBERT on Flickr30K Entities grounding} 
\label{table:comparison}
\end{table}

Finally, we compare the results with the recent work in progress, VisualBERT \citep{Li:2019wy}, in Table~\ref{table:comparison} which also achieves improved grounding results based on a single transformer architecture that learns the representations by fusing text and image inputs in the beginning.
Marginally, ours performs better in the top 1 recall.
Note, our approach, unlike VisualBERT which requires task-agnostic and task-specific pre-training on COCO captioning \citep{Chen:2015ur} and the target dataset, needs no similar pre-training to deliver competitive results.
Besides, our architecture is also flexible to adapt to different input modalities respectively.

\section{Conclusion}
\label{sec:conclusion}

This paper introduces contextual grounding, a higher-order interaction technique to capture corresponding context between text entities and visual objects.
The evaluation shows the SOTA 71.36\% accuracy of phrase localization on Flickr30K Entities.
In the future, it would be worth investigating the benefits of grounding guided visual representations in other related and spatio-temporal tasks.

\clearpage
{\small
\bibliographystyle{plainnat}
\bibliography{ref}

\begin{thebibliography}{14}
\providecommand{\natexlab}[1]{#1}
\providecommand{\url}[1]{\texttt{#1}}
\expandafter\ifx\csname urlstyle\endcsname\relax
  \providecommand{\doi}[1]{doi: #1}\else
  \providecommand{\doi}{doi: \begingroup \urlstyle{rm}\Url}\fi

\bibitem[Anderson et~al.(2018)Anderson, He, Buehler, Teney, Johnson, Gould, and
  Zhang]{Anderson:2018ue}
Peter Anderson, Xiaodong He, Chris Buehler, Damien Teney, Mark Johnson, Stephen
  Gould, and Lei Zhang.
\newblock {Bottom-Up and Top-Down Attention for Image Captioning and Visual
  Question Answering}.
\newblock In \emph{IEEE Conference on Computer Vision and Pattern Recognition},
  2018.

\bibitem[Chen et~al.(2015)Chen, Fang, Lin, Vedantam, Gupta, Doll{\'a}r, and
  Zitnick]{Chen:2015ur}
Xinlei Chen, Hao Fang, Tsung-Yi Lin, Ramakrishna Vedantam, Saurabh Gupta, Piotr
  Doll{\'a}r, and C~Lawrence Zitnick.
\newblock {Microsoft COCO Captions: Data Collection and Evaluation Server}.
\newblock \emph{arXiv}, April 2015.

\bibitem[Devlin et~al.(2019)Devlin, Chang, Lee, and Toutanova]{Devlin:2018uk}
Jacob Devlin, Ming-Wei Chang, Kenton Lee, and Kristina Toutanova.
\newblock {BERT: Pre-training of Deep Bidirectional Transformers for Language
  Understanding.}
\newblock In \emph{Annual Meeting of the Association for Computational
  Linguistics}, 2019.

\bibitem[Girdhar et~al.(2019)Girdhar, Carreira, Doersch, and
  Zisserman]{Girdhar:2019uo}
Rohit Girdhar, Joao Carreira, Carl Doersch, and Andrew Zisserman.
\newblock {Video Action Transformer Network}.
\newblock In \emph{IEEE Conference on Computer Vision and Pattern Recognition},
  2019.

\bibitem[Hinami and Satoh(2017)]{Hinami:2017uz}
Ryota Hinami and Shin'ichi Satoh.
\newblock {Discriminative Learning of Open-Vocabulary Object Retrieval and
  Localization by Negative Phrase Augmentation}.
\newblock In \emph{Conference on Empirical Methods on Natural Language
  Processing}, November 2017.

\bibitem[Kim et~al.(2018)Kim, Jun, and Zhang]{Kim:2018wu}
Jin-Hwa Kim, Jaehyun Jun, and Byoung-Tak Zhang.
\newblock {Bilinear Attention Networks}.
\newblock \emph{IEEE International Conference on Image Processing}, May 2018.

\bibitem[Li et~al.(2019)Li, Yatskar, Yin, Hsieh, and Chang]{Li:2019wy}
Liunian~Harold Li, Mark Yatskar, Da~Yin, Cho-Jui Hsieh, and Kai-Wei Chang.
\newblock {VisualBERT: A Simple and Performant Baseline for Vision and
  Language}.
\newblock \emph{arXiv}, August 2019.

\bibitem[Lin et~al.(2017)Lin, Doll{\'a}r, Girshick, He, Hariharan, and
  Belongie]{Lin:2017vv}
Tsung-Yi Lin, Piotr Doll{\'a}r, Ross Girshick, Kaiming He, Bharath Hariharan,
  and Serge Belongie.
\newblock {Feature pyramid networks for object detection}.
\newblock In \emph{IEEE Conference on Computer Vision and Pattern Recognition},
  2017.

\bibitem[Liu et~al.(2017)Liu, Mao, Sha, and Yuille]{Liu:2017ve}
Chenxi Liu, Junhua Mao, Fei Sha, and Alan~L Yuille.
\newblock {Attention Correctness in Neural Image Captioning.}
\newblock In \emph{AAAI Spring Symposium}, 2017.

\bibitem[Parmar et~al.(2018)Parmar, Vaswani, Uszkoreit, Kaiser, Shazeer, Ku,
  and Tran]{Parmar:2018vc}
Niki Parmar, Ashish Vaswani, Jakob Uszkoreit, {\L}ukasz Kaiser, Noam Shazeer,
  Alexander Ku, and Dustin Tran.
\newblock {Image Transformer}.
\newblock In \emph{International Conference on Machine Learning}, February
  2018.

\bibitem[Plummer et~al.(2015)Plummer, Wang, Cervantes, Caicedo, Hockenmaier,
  and Lazebnik]{Plummer:2015ve}
Bryan~A Plummer, Liwei Wang, Chris~M Cervantes, Juan~C Caicedo, Julia
  Hockenmaier, and Svetlana Lazebnik.
\newblock {Flickr30k Entities: Collecting Region-to-Phrase Correspondences for
  Richer Image-to-Sentence Models}.
\newblock \emph{International Journal of Computer Vision}, May 2015.

\bibitem[Vaswani et~al.(2017)Vaswani, Shazeer, Parmar, Uszkoreit, Jones, Gomez,
  Kaiser, and Polosukhin]{Vaswani:2017ul}
Ashish Vaswani, Noam Shazeer, Niki Parmar, Jakob Uszkoreit, Llion Jones,
  Aidan~N Gomez, Lukasz Kaiser, and Illia Polosukhin.
\newblock {Attention is All you Need}.
\newblock In \emph{Neural Information Processing Systems}, 2017.

\bibitem[Wang et~al.(2018)Wang, Girshick, Gupta, and He]{Wang:2018ut}
Xiaolong Wang, Ross Girshick, Abhinav Gupta, and Kaiming He.
\newblock {Non-Local Neural Networks}.
\newblock In \emph{IEEE Conference on Computer Vision and Pattern Recognition},
  pages 7794--7803, 2018.

\bibitem[Yeh et~al.(2017)Yeh, Xiong, Hwu, Do, and Schwing]{Yeh:2017td}
Raymond Yeh, Jinjun Xiong, Wen-Mei Hwu, Minh Do, and Alexander Schwing.
\newblock {Interpretable and Globally Optimal Prediction for Textual Grounding
  using Image Concepts}.
\newblock In \emph{Neural Information Processing Systems}, pages 1912--1922,
  2017.

\end{thebibliography}
}

\clearpage
\section*{Supplementary Materials}

\subsection*{Implementation}
\label{sec:implementation}

Our implementation is based on \href{https://github.com/pytorch/pytorch/tree/v1.1.0}{PyTorch-v1.1} and \href{https://github.com/huggingface/transformers/tree/v0.6.2}{PyTorch Pretrained BERT-v0.6.2}.
We follow the same training protocol as BAN \citep{Kim:2018wu} on the Flickr30K Entities dataset for fair comparison where the text entity representation is taken from the last word or subword in a phrase.
It probably makes little sense for VisualBERT \citep{Li:2019wy} to choose the cross entropy to rank the correspondences instead of the binary cross entropy because one text entity such as a group of people can actually correspond to multiple person objects in the ground truth annotations.
Apart from different number of transformer layers and attention heads used in the base BERT model, our image transformer branch uses gradient clipping of 0.25 and dropout probability 0.4 for the best performance.
During training, the learning rate is set to 5e-5 and the batch size is set to 256 with 2 steps of gradient accumulation before back-propagation.
The model is trained for at most 10 epochs with early stopping.

\end{document}